\documentclass[runningheads]{llncs}
\usepackage{graphicx}
\usepackage{comment}
\usepackage{amsmath,amssymb} 
\usepackage{color}
\usepackage{multirow}
\usepackage{hyperref}

\begin{document}
\pagestyle{headings}
\mainmatter
\def\ECCVSubNumber{978}  

\title{Unpaired Image-to-Image Translation using Adversarial Consistency Loss} 

\titlerunning{ACL-GAN}
%
\author{Yihao Zhao \inst{1}\and
Ruihai Wu \inst{1}\and
Hao Dong \inst{1,2}\thanks{corresponding author}}
%
%
\institute{
	Hyperplane Lab, CFCS, Computer Science Dept., Peking University \\
	\and
	Peng Cheng Lab\\
\email{\{zhaoyh98, wuruihai, hao.dong\}@pku.edu.cn}}
\maketitle

\begin{abstract}
Unpaired image-to-image translation is a class of vision problems whose goal is to find the mapping between different image domains using unpaired training data. 
Cycle-consistency loss is a widely used constraint for such problems. However, due to the strict pixel-level constraint, it cannot perform shape changes, remove large objects, or ignore irrelevant texture.
In this paper, we propose a novel adversarial-consistency loss for image-to-image translation. 
This loss does not require the translated image to be translated back to be a specific source image but can encourage the translated images to retain important features of the source images and overcome the drawbacks of cycle-consistency loss noted above.
Our method achieves state-of-the-art results on three challenging tasks: glasses removal, male-to-female translation, and selfie-to-anime translation.
The code is available at \href{https://github.com/hyperplane-lab/ACL-GAN}{https://github.com/hyperplane-lab/ACL-GAN}
\end{abstract}

\section{Introduction}
\label{introduction}

\begin{figure}[h!]
    \begin{center}
        \includegraphics[scale=0.465, trim={1.7cm, 8.05cm, 1.7cm, 7.65cm}, clip]{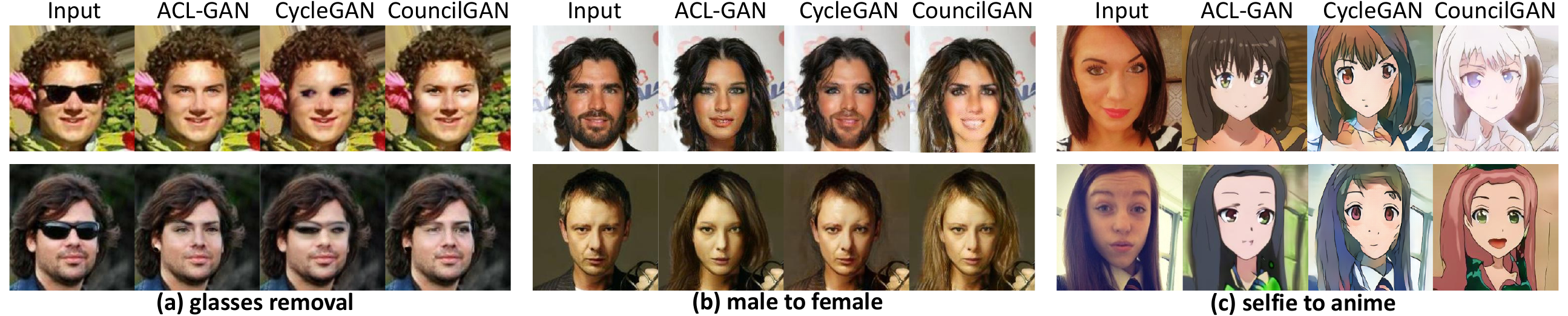}
    \end{center}
    \caption{\textbf{Example results of our ACL-GAN and baselines.} 
    Our method does not require cycle consistency, so it can bypass unnecessary features. Moreover, with the proposed adversarial-consistency loss, our method can explicitly encourage the generator to maintain the commonalities between the source and target domains.
    }
    \label{fig:pic1}
\end{figure}

Learning to translate data from one domain to another is an important problem in computer vision for its wide range of applications, such as 
colourisation~\cite{zhang2016colorful}, super-resolution~\cite{ChaoTPAMI2016,Kim2016CVPR}, and image inpainting~\cite{pathakCVPR16context,IizukaSIGGRAPH2017}. 
In unsupervised settings where only unpaired data are available, current methods~\cite{annosheh2018cvprw,choi2018stargan,taigman2017unsupervised,huang2018munit,kim2017discogan,liu2017unsupervised,DBLP:journals/corr/abs-1711-05139,yi2017dualgan,zhu2017cyclegan} mainly rely on shared latent space and the assumption of cycle-consistency.
In particular, by forcing the translated images to fool the discriminator using a classical adversarial loss and translating those images back to the original images, cycle-consistency ensures the translated images contain enough information from the original input images. This helps to build a reasonable mapping and generate high-quality results.

The main problem of cycle-consistency is that it assumes the translated images contain all the information of the input images in order to reconstruct the input images. This assumption leads to the preservation of source domain features, such as traces of large objects (\emph{e.g.,} the trace of glasses in Fig.~\ref{fig:pic1} $(a)$) and texture (\emph{e.g.,} the beard remains on the face in Fig.~\ref{fig:pic1} $(b)$), resulting in unrealistic results. Besides, the cycle-consistency constrains shape changes of source images (\emph{e.g.,} the hair and the shape of the face are changed little in Fig.~\ref{fig:pic1} $(b)$ and $(c)$).

To avoid the drawbacks of cycle-consistency, we propose a novel method, termed ACL-GAN, where ACL stands for \emph{adversarial-consistency loss}.
Our goal is to maximise the information shared by the target domain and the source domain.
The adversarial-consistency loss encourages the generated image to include more features from the source image, from the perspective of distribution, rather than maintaining the pixel-level consistency. 
That's to say, the image translated back is only required to be similar to the input source image but need not be identical to a specific source image.
Therefore, our generator is not 
required to preserve all the information from the source image, which avoids leaving artefacts in the translated image.
Combining the adversarial-consistency loss with classical adversarial loss and some additional losses makes our full objective for unpaired image-to-image translation.


The main contribution of our work is a novel method to support unpaired image-to-image translation which avoids the drawbacks of cycle-consistency. Our method achieves state-of-the-art results 
on three challenging tasks: glasses removal, male-to-female translation, and selfie-to-anime translation.

\section{Related Work}
\label{related_work}

Generative adversarial networks (GANs)~\cite{goodfellow2014gan} have been successfully applied to numerous image applications, such as super-resolution~\cite{Ledig2017photo} and image colourisation~\cite{zhang2016colorful}. These tasks can be seen as ``translating'' an image into another image. Those two images usually belong to two different domains, termed source domain and target domain.
Traditionally, each of the image translation tasks is solved by a task-specific method. To achieve task-agnostic image translation, Pix2Pix~\cite{Isola2017} was the first work to support different image-to-image translation tasks using a single method.
However, it requires paired images to supervise the training.

Unpaired image-to-image translation has gained a great deal of attention for applications in which paired data are unavailable or difficult to collect.
A key problem of unpaired image-to-image translation is determining which properties of the source domain to preserve in the translated domain, and how to preserve them.
Some methods have been proposed to preserve pixel-level properties, such as pixel gradients~\cite{bousmalis2017pixel}, pixel values~\cite{shrivastava2017pixel}, and pairwise sample distances~\cite{benaim2017pairwise}.
Recently, several concurrent works, CycleGAN~\cite{zhu2017cyclegan}, DualGAN~\cite{yi2017dualgan}, DiscoGAN~\cite{kim2017discogan}, and UNIT~\cite{liu2017unsupervised}, achieve unpaired image-to-image translation using cycle-consistency loss as a constraint.
Several other studies improve image-to-image translation in different aspects.
For example, BicycleGAN~\cite{zhu2017toward} supports multi-modal translation using paired images as the supervision, while Augmented CycleGAN~\cite{Almahairi2018augmentedCycleGAN} achieves that with unpaired data.
DRIT~\cite{Lee2018DRIT,Lee2019DRITplus} and MUNIT~\cite{huang2018munit} disentangle domain-specific features and support multi-modal translations.
To improve image quality, attention CycleGAN~\cite{Mejjati2018attentionCycleGAN} proposes an attention mechanism to preserve the background from the source domain.
In this paper, we adopt the network architecture of MUNIT but we do not use cycle-consistency loss.

To alleviate the problem of cycle-consistency, 
a recent work, CouncilGAN~\cite{Nizan2019breackcycle}, has proposed using duplicate generators and discriminators together with the council loss as a replacement for cycle-consistency loss.
By letting different generators compromise each other, the council loss encourages the translated images to contain more information from the source images.
By contrast, our method does not require duplicate generators and discriminators, so it uses fewer network parameters and needs less computation. 
Moreover, our method can generate images of higher quality than CouncilGAN~\cite{Nizan2019breackcycle}, because our adversarial-consistency loss explicitly makes the generated image similar to the source image, while CouncilGAN~\cite{Nizan2019breackcycle} only compromises among generated images.
Details are described in Section~\ref{ACL_LOSS}.

\section{Method}
\label{method}

Our goal is to translate images from one domain to the other domain and support diverse and multi-modal outputs.
Let $X_{S}$ and $X_{T}$ be the source and target domains, $X$ be the union set of $X_S$ and $X_T$ (\emph{i.e.,} $X = X_S \cup X_T$), $x \in X$ be a single image, $x_{S} \in X_{S}$ and $x_{T} \in X_{T}$ be the images of different domains. We define $p_X$, $p_{S}$ and $p_{T}$ to be the distributions of $X$, $X_S$ and $X_T$. $p_{(a,b)}$ is used for joint distribution of pair $(a,b)$, where $a$ and $b$ can be images or noise vectors. Let $Z$ be the noise vector space, $z\in Z$ be a noise vector and  $z\sim \mathcal{N}(0,1)$.

Our method has two generators: $G_{S}: (x, z) \rightarrow x_{S}$ and $G_{T}: (x, z) \rightarrow x_{T}$ which translate images to domain $X_S$ and $X_T$, respectively. 
Similar to~\cite{huang2018munit}, each generator contains a noise encoder, an image encoder, and a decoder, in which the noise encoder is used only for calculating the identity loss. 
The generators receive input pairs (image, noise vector) where the image is from $X$.
In detail, the image encoder receives images sampled from $X$.
The noise vector $z$ obtained from the noise encoder is only for identity loss, while for other losses, the noise vector $z$ is randomly sampled from the standard normal distribution, $\mathcal{N}(0,1)$.
The noise vector $z$ and the output of the image encoder are forwarded to the decoder and the output of the decoder is the translated image.

Moreover, there are two kinds of discriminators $D_S$/$D_T$ and $\hat{D}$. 
$\hat{D}$ is a consistency discriminator. Its goal is to ensure the consistency between source images and translated images, and this is the core of our method.
The goal of $D_S$ and $D_T$ is to distinguish between real and fake images in a certain domain. Specifically, the task of $D_S$ is to distinguish between $X_S$ and $G_S(X)$, and the task of $D_T$ is to distinguish between $X_T$ and $G_T(X)$.

\begin{figure}[t]
    \begin{center}
        \includegraphics[scale=0.5, trim={4cm, 8.3cm, 4cm, 4cm}, clip]{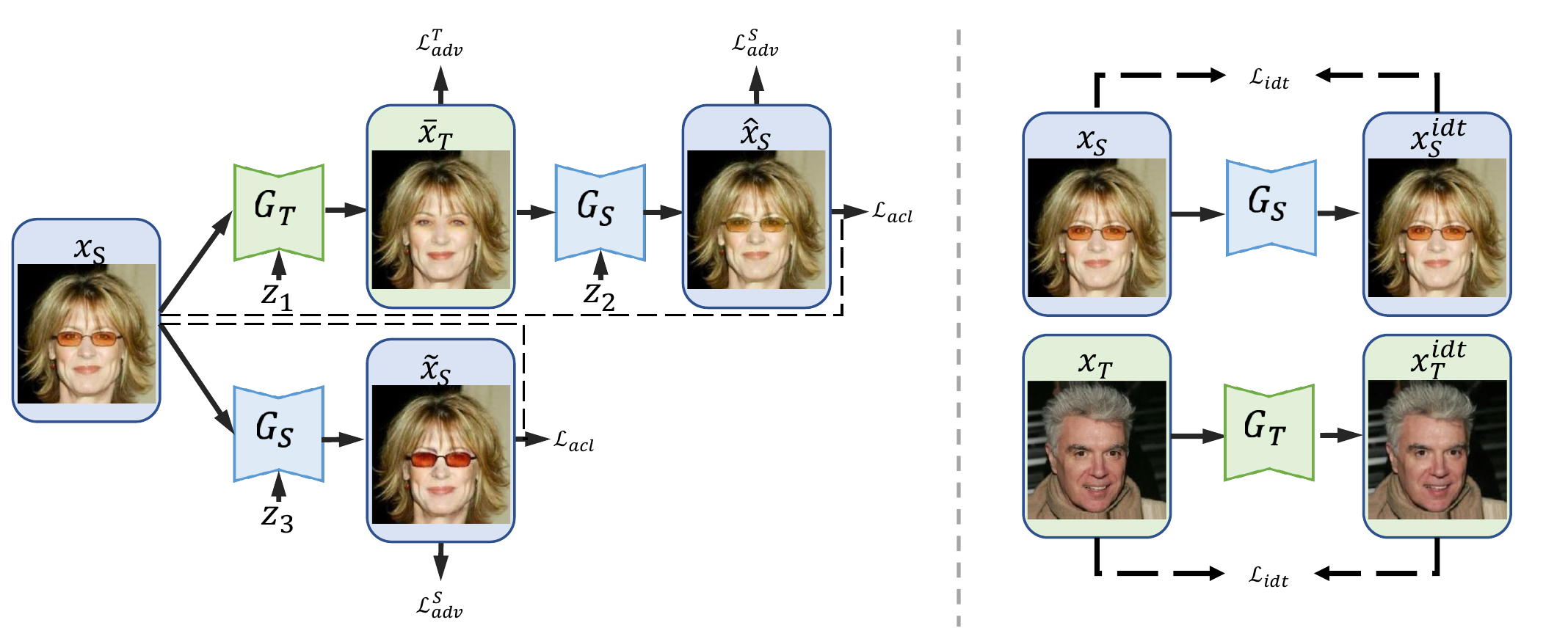}
    \end{center}
    \caption{\textbf{The training schema of our model.}
    (Left) Our model contains two generators: $G_S: (X, Z) \rightarrow X_S$ and $G_T: (X, Z) \rightarrow X_T$ and three discriminators: $D_S, D_T$ for $\mathcal{L}_{adv}^S, \mathcal{L}_{adv}^T$ and $\hat{D}$ for $\mathcal{L}_{acl}$. $D_S$ and $D_T$ ensure that the translated images belong to the correct image domain, while $\hat{D}$ encourages the translated images to preserve important features of the source images. The noise vectors $z_1, z_2, z_3$ are randomly sampled from $\mathcal{N}(0,1)$.
    (Right) $\mathcal{L}_{idt}$ encourages to maintain features, improves the image quality, stabilises the training process and prevents mode collapse, where the noise vector is from the noise encoder.
    The blocks with the same colour indicate shared parameters.
    }
    \label{fig:overall_model}
\end{figure}

The objective of ACL-GAN has three parts.
The first, \emph{adversarial-translation loss}, matches the distributions of generated images to the data distributions in the target domain.
The second, \emph{adversarial-consistency loss}, preserves significant features of the source images in the translated images, \emph{i.e.,} it results in reasonable mappings between domains.
The third, identity loss and bounded focus mask, can further help to improve the image quality and maintain the image background.
The data are forwarded as shown in Fig.~\ref{fig:overall_model}, and the details of our method are described below.

\subsection{Adversarial-Translation Loss}
For image translation between domains, we utilise 
the classical adversarial loss, which we call \emph{adversarial-translation loss} in our method, to both generators, $G_S$ and $G_T$, and discriminators, $D_S$ and $D_T$. For generator $G_T$ and its discriminator $D_T$, the adversarial-translation loss is as follows:
\begin{equation}
\begin{aligned}
\mathcal{L}^T_{adv}(G_T, D_T, X_S, X_T) &= \mathbb{E}_{x_T \sim p_{T}} [log D_T(x_T)] 
\\&
+\mathbb{E}_{\bar{x}_T \sim p_{\{\bar{x}_T\}}}[log (1-D_T(\bar{x}_T))]
\end{aligned}
\label{equi:loss_adv_T}
\end{equation}
where $\bar{x}_T = G_T(x_S, z_1)$ and $z_1\sim \mathcal{N}(0,1)$.
The objective is $min_{G_T} max_{D_T}$
$\mathcal{L}^T_{adv}(G_T, D_T, X_S, X_T)$.

The discriminator $D_S$ is expected to distinguish between real images of domain $X_S$ and translated images generated by $G_S$. 
The generator $G_S$ tries to generate images, $\hat{x}_S$ and $\tilde{x}_S$, that look similar to images from domain $X_S$.
Therefore, the loss function is defined as: 
\begin{equation}
\begin{aligned}
\mathcal{L}^S_{adv}(G_S, D_S, \{\bar{x}_T\}, X_S) &= \mathbb{E}_{x_S \sim p_S} [log D_S(x_S)] \\
&+(\mathbb{E}_{\hat{x}_S \sim p_{\{\hat{x}_S\}}}[log (1-D_S(\hat{x}_S))] \\
&+\mathbb{E}_{\tilde{x}_S \sim p_{\{\tilde{x}_S\}}}[log (1-D_S(\tilde{x}_S))])/2
\end{aligned}
\label{equi:loss_adv_S}
\end{equation}
where $\hat{x}_S = G_S(\bar{x}_T, z_2)$, $\tilde{x}_S = G_S(x_S, z_3), z_2, z_3\sim \mathcal{N}(0,1)$ and the objective is $min_{G_S} max_{D_S}\mathcal{L}^S_{adv}(G_S, D_S, \{\bar{x}_T\}, X_S)$.
In summary, we can define the adversarial-translation loss as follows:
\begin{equation}
\begin{aligned}
\mathcal{L}_{adv} &= \mathcal{L}^T_{adv}(G_T, D_T, X_S, X_T) + \mathcal{L}^S_{adv}(G_S, D_S, \{\bar{x}_T\}, X_S)
\end{aligned}
\label{equi:loss_adv}
\end{equation}

\subsection{Adversarial-Consistency Loss} 
\label{ACL_LOSS}




\begin{figure}[t]
    \begin{center}
        \includegraphics[scale=0.58, trim={5.3cm, 7.6cm, 4cm, 8cm}, clip]{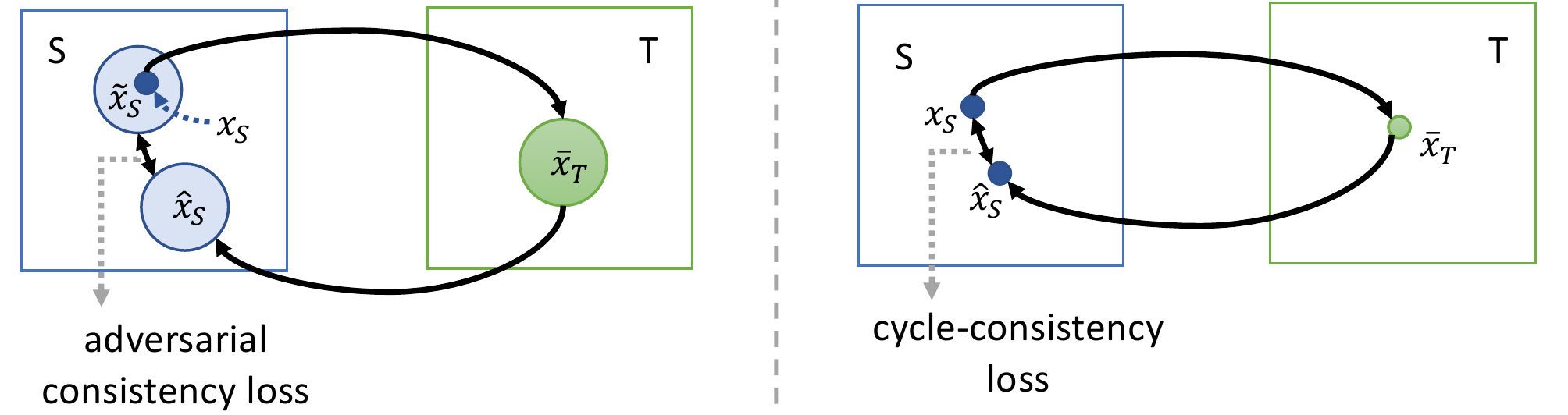}
    \end{center}
    \caption{\textbf{The comparison of adversarial-consistency loss and cycle-consistency loss~\cite{zhu2017cyclegan}.}
    The blue and green rectangles represent image domains S and T, respectively. 
    Any point inside a rectangle represents a specific image in that domain.
    (Right): given a source image $x_S$, cycle-consistency loss requires the image translated back, $\hat{x}_S$, should be identical to the source image, $x_S$. 
    (Left): given a source image $x_S$, we synthesise multi-modal images in its neighbourhood distribution $\tilde{x}_S$, the distribution $\bar{x}_T$ in the target domain $X_T$, and the distribution $\hat{x}_S$ translated from $\bar{x}_T$. The distributions are indicated by the blue and green circles.
    Instead of requiring the image translated back, $\hat{x}_S$, to be a specific image, we minimise the distance between the distributions of $\hat{x}_S$ and $\tilde{x}_S$, so that a specific $\hat{x}_S$ can be any point around $x_S$. 
    By doing so, we encourage $\bar{x}_T$ to preserve the features of the original image $x_S$.
    }
    \label{fig:loss_cmp}
\end{figure}

The $\mathcal{L}_{adv}$ loss described above can encourage the translated image, $\bar{x}_T$, to be in the correct domain $X_T$. However, this loss cannot encourage the translated image $\bar{x}_T$ to be similar to the source image $x_S$. For example, when translating a male to a female, the facial features of the female might not be related to those of the male.
%
To preserve important features of the source image in the translated image, we propose the adversarial-consistency loss, which is realised by a consistency discriminator $\hat{D}$.
The consistency discriminator impels the generator to minimise the distance between images $\tilde{x}_S$ and $\hat{x}_S$ as shown in Fig.~\ref{fig:loss_cmp}.
The "real" and "fake" images for $\hat{D}$ can be swapped without affecting performance.
However, letting $\hat{D}$ only distinguish $\hat{x}_S$ and $\tilde{x}_S$ does not satisfy our needs, because the translated images $\hat{x}_S$ and $\tilde{x}_S$ need only to belong to domain $X_S$; they are not required to be close to a specific source image. 
Therefore, the consistency discriminator $\hat{D}$ uses $x_S$ as a reference and adopts paired images as inputs to let the generator minimise the distances between the joint distributions of $(x_S, \hat{x}_S)$ and $(x_S, \tilde{x}_S)$.
In this way, the consistency discriminator $\hat{D}$ encourages the image translated back, $\hat{x}_S$, to contain the features of the source image $x_S$. 
As $\hat{x}_S$ is generated from $\bar{x}_T$, this can encourage the translated image $\bar{x}_T$ to preserve the features of the source image $x_S$.



The input noise vector $z$ enables multi-modal outputs, which is essential to make our method work. 
Without multi-modal outputs, given a specific input image $x_S$, the $\tilde{x}_S$ can have only one case. Therefore, mapping $(x_S, \hat{x}_S)$ and $(x_S, \tilde{x}_S)$ together is almost equivalent to requiring $\hat{x}_S$ and $x_S$ to be identical. This strong constraint is similar to cycle-consistent loss whose drawbacks have been discussed before.
%
With the multi-modal outputs, given a specific image $x_S$, the $\tilde{x}_S$ can have many possible cases. Therefore, the consistency discriminator $\hat{D}$ can focus on the feature level, rather than the pixel level. That is to say, $\hat{x}_S$ does not have to be identical to a specific image $x_S$. 
For example, when translating faces with glasses to faces without glasses, the $\tilde{x}_S$ and $\hat{x}_S$ can be faces with different glasses, {\emph{e.g.}} the glasses have different colours and frames in $x_S$, $\tilde{x}_S$ and $\hat{x}_S$ in Fig.~\ref{fig:overall_model}.
Thus, $\bar{x}_T$ need not retain any trace of glasses at the risk of increasing $\mathcal{L}_{adv}^{T}$, and $\mathcal{L}_{acl}$ can still be small.



The adversarial-consistency loss (ACL) is as follows:
\begin{equation}
\begin{aligned}
\mathcal{L}_{acl} &= \mathbb{E}_{(x_S, \hat{x}_S) \sim p_{(X_S, \{\hat{x}_S\})}}[log \hat{D}(x_S, \hat{x}_S)]\\
&+ \mathbb{E}_{(x_S, \tilde{x}_S) \sim p_{(X_S, \{\tilde{x}_S\})}}[log (1 - \hat{D}(x_S, \tilde{x}_S))]
\end{aligned}
\label{equi:loss_acl}
\end{equation}
where $x_S \in X_S$, $\bar{x}_T=G_T(x_S, z_1)$, $\hat{x}_S=G_S(\bar{x}_T, z_2)$, $\tilde{x}_S=G_S(x_S, z_3)$.

\subsection{Other Losses}
\label{OTHER_LOSS}

\textbf{Identity loss} 
We further apply identity loss to encourage the generators to be approximate identity mappings when images of the target domain are given to the generators.
Identity loss can further encourage feature preservation, improve the quality of translated images, stabilise the training process, and avoid mode collapse because 
the generator is required to be able to synthesise all images in the dataset~\cite{zhu2017cyclegan,rosca2017alphagan}. 
Moreover, the identity loss between the source image $x_S$ and the reconstruction image $x^{idt}_S$ can guarantee that $x_S$ is inside the distribution of $\tilde{x}_S$ as shown in Fig.~\ref{fig:loss_cmp}. 

We formalise two noise encoder networks, $E^z_S: X_S \rightarrow Z$ and $E^z_T: X_T \rightarrow Z$ for $G_S$ and $G_T$, respectively, which map the images to the noise vectors. 
Identity loss can be formalised as: 
\begin{equation}
\begin{aligned}
\mathcal{L}_{idt} = \mathbb{E}_{x_S \sim p_S} [|| x_S - x_S^{idt} ||_1] + \mathbb{E}_{x_T \sim p_T} [|| x_T - x_T^{idt} ||_1]
\end{aligned}
\label{equi:loss_idt}
\end{equation}
where $x_S^{idt} = G_S(x_S, E^z_S(x_S))$ and $x_T^{idt} = G_T(x_T, E^z_T(x_T))$.

\textbf{Bounded focus mask} Some applications require the generator to only modify certain areas of the source image and keep the rest unchanged. 
We let the generator produce four channels, where the first three are the channels of RGB images and the fourth is called bounded focus mask whose values are between 0 and 1.
The translated image $x_T$ can be obtained by the formula: $x_T = x\prime_T \odot x_m + x_S \odot (1-x_m)$, where $\odot$ is element-wise product, $x_S$ is the source image, $x\prime_T$ is the first three output channels of the generator and $x_m$ is the bounded focus mask. 
We add the following constraints to the generator which is one of our contributions: 
\begin{equation}
\begin{aligned}
\mathcal{L}_{mask} &= \delta[(\max \{\sum_k x_m[k] - \delta_{max}\times W, 0\})^2\\
& + (\max \{\delta_{min}\times W - \sum_k x_m[k], 0\})^2]\\
& + \sum_k \frac{1}{|x_m[k]-0.5|+\epsilon}
\end{aligned}
\label{equi:loss_focus}
\end{equation}
where $\delta$, $\delta_{max}$ and $\delta_{min}$ are hyper-parameters for controlling the size of masks, $x_m[k]$ is the k-th pixel of the mask and $W$ is the number of pixels of an image. 
The $\epsilon$ is a marginal value to avoid dividing by zero.
The first term of this loss limits the size of the mask to a suitable range. It encourages the generator to make enough changes and maintain the background, where $\delta_{max}$ and $\delta_{min}$ are the maximum and minimum proportions of the foreground in the mask.
The minimum proportion is essential for our method because it avoids $\tilde{x}_S$ being identical to $x_S$ under different noise vectors.
The last term of this loss encourages the mask values to be either 0 or 1 to segment the image into a foreground and a background~\cite{Nizan2019breackcycle}.
In the end, this loss is normalised by the size of the image. 

\subsection{Implementation Details}
\textbf{Full objective.} Our total loss is as follows:
\begin{equation}
\begin{aligned}
\mathcal{L}_{total} &= \mathcal{L}_{adv} + \lambda_{acl}\mathcal{L}_{acl} 
+ \lambda_{idt}\mathcal{L}_{idt} + \lambda_{mask}\mathcal{L}_{mask}
\end{aligned}
\label{equi:loss_total}
\end{equation}
where $\lambda_{acl}, \lambda_{idt}, \lambda_{mask}$ are all scale values that control the weights of different losses.
We compare our method against ablations of the full objective in Section~\ref{ablation_studyies} to show the importance of each component. 

\textbf{Network architecture} 
For generator and discriminator, we follow the design of those in~\cite{huang2018munit}. Specifically, a generator consists of two encoders and one decoder.
Besides the image encoder and decoder that together form an auto-encoder architecture, our model employs the noise encoder, whose architecture is similar to the style encoder in~\cite{huang2018munit}.
Meanwhile, our discriminators use a multi-scale technique~\cite{wang2018scale} to improve the visual quality of synthesised images.

\textbf{Training details}
We adopt a least-square loss~\cite{mao2017lsgan} for $\mathcal{L}_{adv}$ (Equation~\ref{equi:loss_adv}) and $\mathcal{L}_{acl}$ (Equation~\ref{equi:loss_acl}). This loss brings more stable training process and better results compared with~\cite{goodfellow2014gan}.
For all the experiments, we used Adam optimiser~\cite{kingma2014adam} with $\beta_1 = 0.5$ and $\beta_2 = 0.999$. The batch size was set to $3$. All models were trained with a learning rate of 0.0001 and the learning rate dropped by a factor of $0.5$ after every $100K$ iterations. We trained all models for $350K$ iterations.
The discriminators update twice while the generators update once.
For training, we set $\delta=0.001$, $\epsilon=0.01$ and $\lambda_{idt}=1$.
The values of $\lambda_{acl}$, $\lambda_{mask}$, $\delta_{min}$ and $\delta_{max}$ were 
set according to different applications: 
in glasses removal $\lambda_{acl}=0.2, \lambda_{mask}=0.025, \delta_{min}=0.05, \delta_{max}=0.1$;
in male-to-female translation $\lambda_{acl}=0.2, \lambda_{mask}=0.025, \delta_{min}=0.3, \delta_{max}=0.5$; 
in selfie-to-anime translation $\lambda_{acl}=0.5, \lambda_{mask}=\delta_{min}=\delta_{max}=0$.
For fair comparison, we follow the same data augmentation method described in CouncilGAN~\cite{Nizan2019breackcycle}.


\section{Experiments}
\label{experiments}

\subsection{Experimental Settings} 

\textbf{Datasets} 
We evaluated ACL-GAN on two different datasets, CelebA~\cite{liu2015celeba} and selfie2anime~\cite{Kim2020U-GAT-IT:}. 
CelebA~\cite{liu2015celeba} contains 202,599 face images with 40 binary attributes. We used the attributes of gender and with/without glasses for evaluation. For both attributes, we used 162,770 images for training and 39,829 images for testing.
For gender, the training set contained 68,261 images of males and 94,509 images of females and the test set contained 16,173 images of males and 23,656 images of females. 
For with/without glasses, the training set contained 10,521 images with glasses and 152,249 images without glasses and the test set contained 2,672 images with glasses and 37,157 images without glasses.
Selfie2anime~\cite{Kim2020U-GAT-IT:} contains 7,000 images. The size of the training set is 3,400 for both anime images and selfie images. The test set has 100 anime images and 100 selfie images.


\textbf{Metrics}
Fr\'echet Inception Distance (FID) ~\cite{heusel2017fid} is an improvement of Inception Score (IS)~\cite{salimans2017is} for evaluating the image quality of generative models.
FID calculates the Fr\'echet distance with mean and covariance between the real and the fake image distributions. 
Kernel Inception Distance (KID)~\cite{binkowski2018mmd} is an improved measure of GAN convergence and quality. It is the squared Maximum Mean Discrepancy between inception representations.

\textbf{Baselines.} 
We compared the results of our method with those of some state-of-the-art models, including CycleGAN~\cite{zhu2017cyclegan}, MUNIT~\cite{huang2018munit}, DRIT++~\cite{Lee2018DRIT,Lee2019DRITplus}, StarGAN~\cite{choi2018stargan}, U-GAT-IT~\cite{Kim2020U-GAT-IT:}, Fixed-Point GAN~\cite{siddiquee2019learning}, and CouncilGAN~\cite{Nizan2019breackcycle}.
All those methods use unpaired training data.
MUNIT~\cite{huang2018munit}, DRIT++~\cite{Lee2018DRIT,Lee2019DRITplus}, and CouncilGAN~\cite{Nizan2019breackcycle} can generate multiple results for a single input image and the others produce only one image for each input.
Among those methods, only CouncilGAN~\cite{Nizan2019breackcycle} does not use cycle-consistency loss. Instead, it adopts $N$ duplicate generators and $2N$ duplicate discriminators ($N$ is set to be 2, 4 or 6 in their paper). This requires much more computation and memory for training.
We set $N$ to be 4 in our comparison.

The quantitative and qualitative results of the baselines were obtained by running the official public codes, except for the CouncilGAN, where the results were from our reproduction.

\begin{figure}[t]
    \begin{center}
        \includegraphics[scale=0.5, trim={3.5cm, 4cm, 2cm, 3.8cm}, clip]{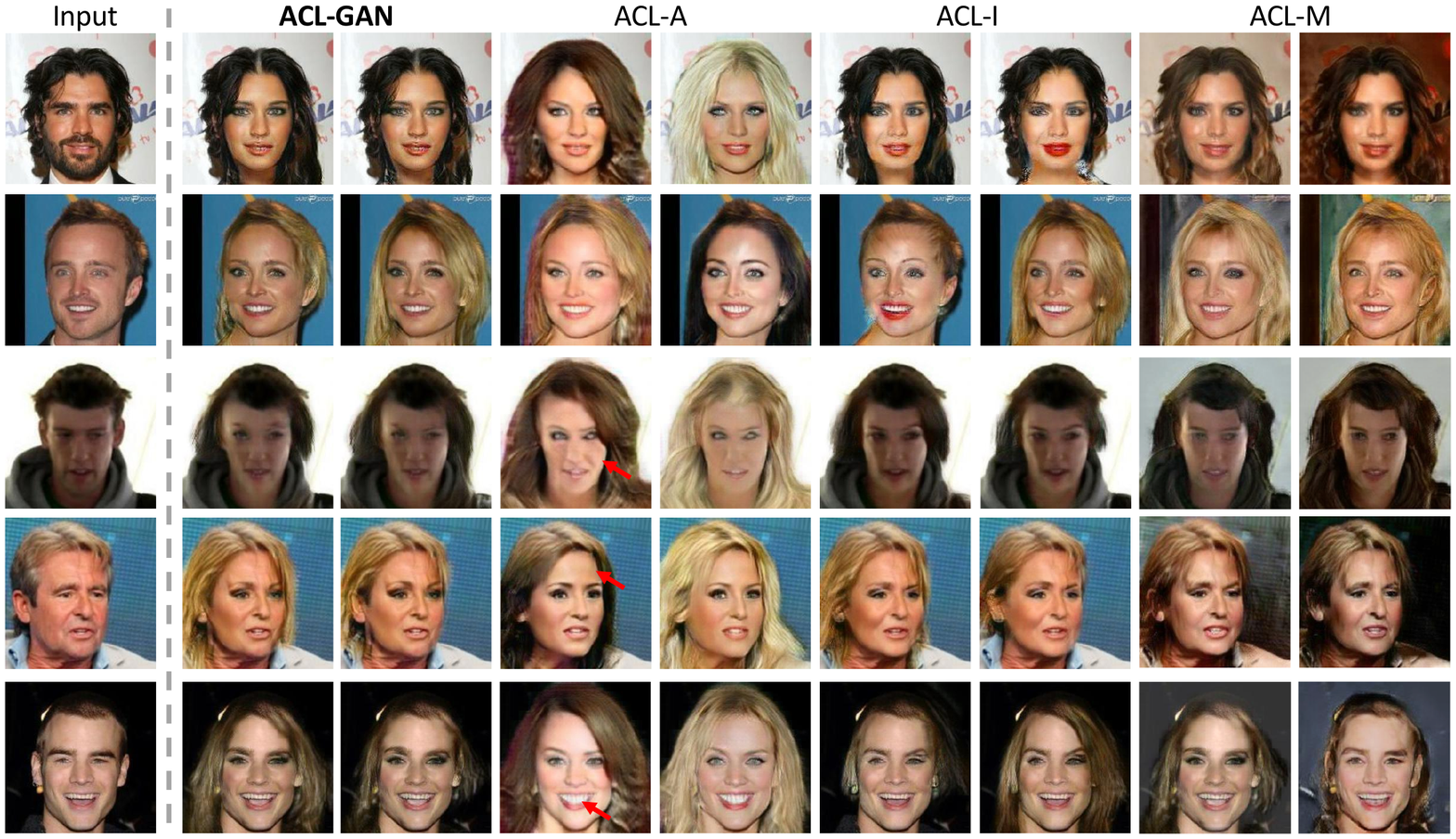}
    \end{center}
    \caption{\textbf{Ablation studies.} 
    The male-to-female translation results illustrate the importance of different losses. From left to right: input images; ACL-GAN (with total loss); ACL-A (without $\mathcal{L}_{acl}$); ACL-I (without $\mathcal{L}_{idt}$); ACL-M (without $\mathcal{L}_{mask}$).}
    \label{fig:ablation}
\end{figure}

\subsection{Ablation Studies}
\label{ablation_studyies}

\begin{table}[t]
\centering
\begin{tabular}{c|c|c|c|c|c}
\hline
\textbf{Model} & \textbf{$\mathcal{L}_{acl}$} & \textbf{$\mathcal{L}_{idt}$} & \textbf{$\mathcal{L}_{mask}$} & \textbf{FID}   & \textbf{KID}                \\ \hline
ACL-A          & -                            & $\checkmark$                 & $\checkmark$                   & 23.96          & 0.023 $\pm$ 0.0003          \\
ACL-I          & $\checkmark$                 & -                            & $\checkmark$                   & 18.39          & 0.018 $\pm$ 0.0004          \\
ACL-M          & $\checkmark$                 & $\checkmark$                 & -                              & 17.72          & 0.017 $\pm$ 0.0004          \\
ACL-GAN        & $\checkmark$                 & $\checkmark$                 & $\checkmark$                   & \textbf{16.63} & \textbf{0.015 $\pm$ 0.0003} \\ \hline
\end{tabular}
\caption{\textbf{Quantitative results of different ablation settings on male-to-female translation.}
For both FID and KID, lower is better. 
For KID, mean and standard deviation are listed.
The ACL-GAN with total loss outperforms other settings.}
\label{table:ablation}
\end{table}

We analysed ACL-GAN by comparing four different settings: 
1) with total loss (ACL-GAN), 
2) without adversarial-consistency loss $\mathcal{L}_{acl}$ ( ACL-A), 
3) without identity loss $\mathcal{L}_{idt}$ (ACL-I), 
and 4) without bounded focus mask and its corresponding loss $\mathcal{L}_{mask}$ (ACL-M).
The results of different settings are shown in Fig.~\ref{fig:ablation} and the quantitative results are listed in Table~\ref{table:ablation}.

Qualitatively, we observed that our adversarial-consistency loss, $\mathcal{L}_{acl}$, successfully helps to preserve important features of the source image in the translated image,
compared with the setting of ACL-A which already has identity loss and bounded focus mask.
For example, as the red arrows on results of ACL-A show, without $\mathcal{L}_{acl}$, the facial features, $\emph{e.g.}$ skin colour, skin wrinkles, and teeth, are difficult to maintain, and the quantitative results are the worst.
We found that ACL-GAN with bounded focus mask has better perceptual quality, \emph{e.g.,} the background can be better maintained, compared with ACL-M.
The quantitative results indicate that our bounded focus mask improves the quality of images because the mask directs the generator to concentrate on essential parts for translation.
Even though it is difficult to qualitatively compare ACL-I with the others, ACL-GAN achieves better quantitative results than ACL-I. 

\begin{figure}[t]
    \begin{center}
        \includegraphics[scale=0.6, trim={3.5cm, 2.7cm, 6cm, 2.7cm}, clip]{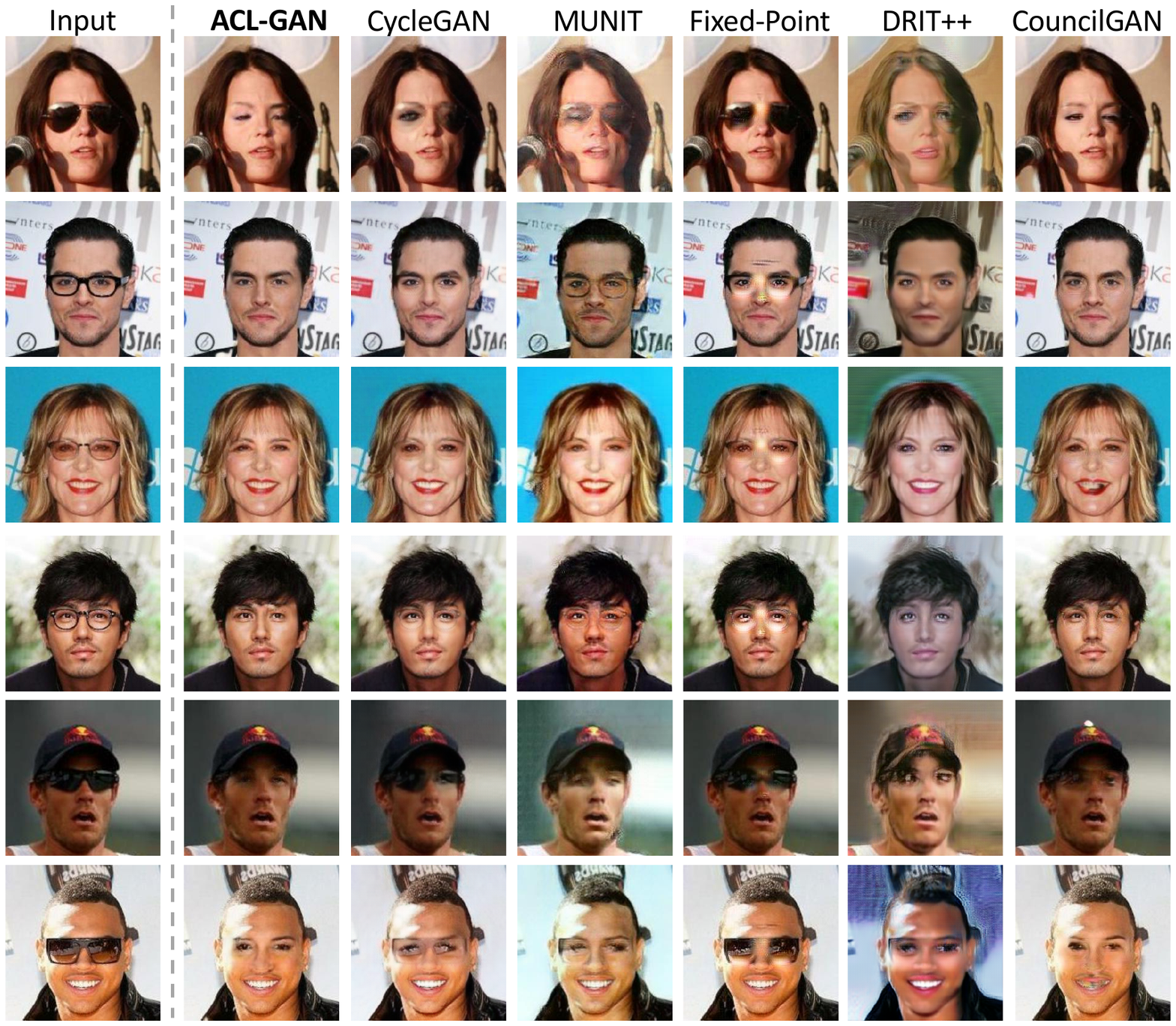}
    \end{center}
    \caption{\textbf{Comparison against baselines on glasses removal.} 
    From left to right: input, our ACL-GAN, CycleGAN~\cite{zhu2017cyclegan}, MUNIT~\cite{huang2018munit}, Fixed-Point GAN~\cite{siddiquee2019learning}, DRIT++~\cite{Lee2018DRIT,Lee2019DRITplus}, and CouncilGAN~\cite{Nizan2019breackcycle}. }
    \label{fig:glasses}
\end{figure}

\begin{figure*}[t]
    \begin{center}
        \includegraphics[scale=0.58, trim={4.5cm, 2.7cm, 4.5cm, 2.6cm}, clip]{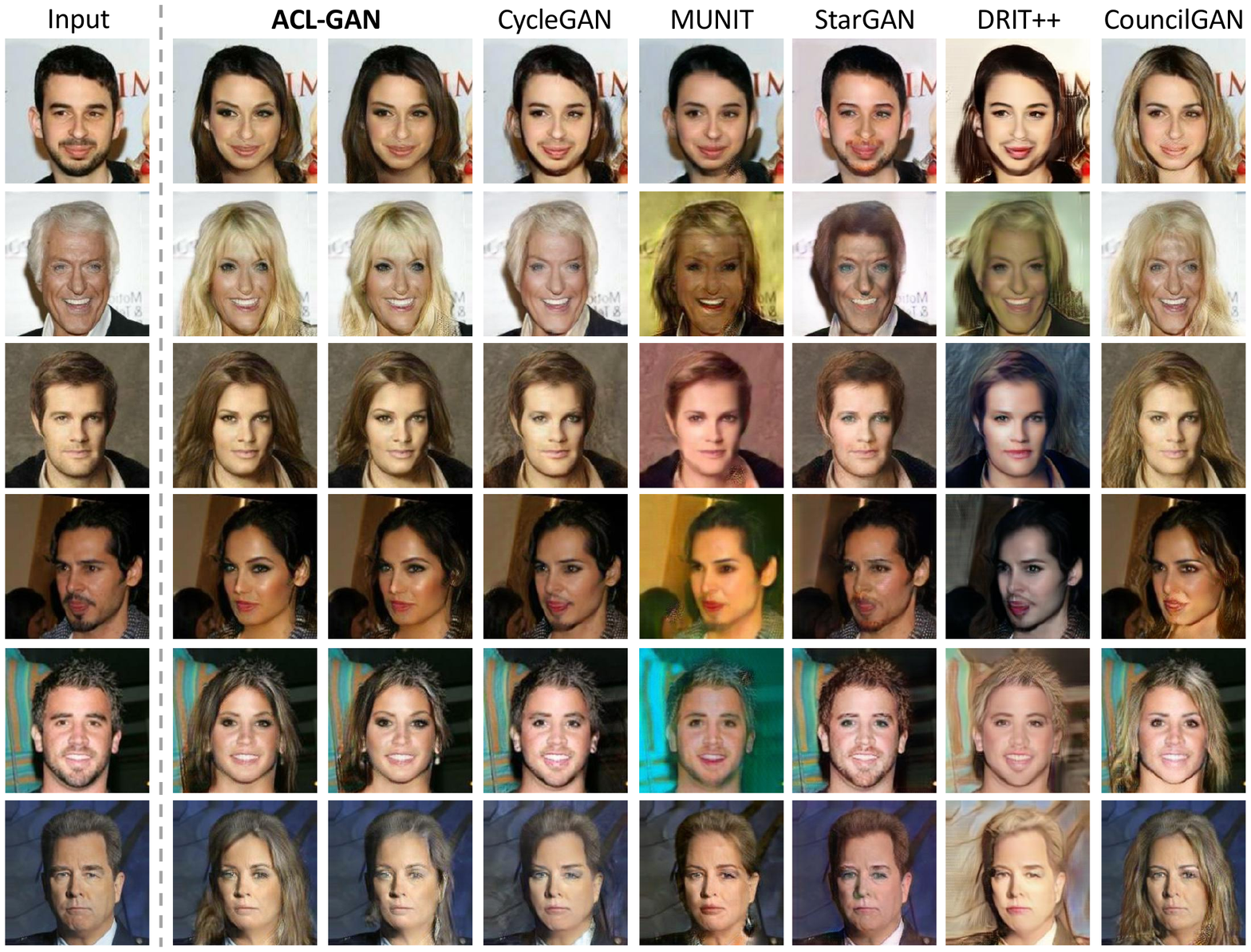}
    \end{center}
    \caption{\textbf{Comparison against baselines on male-to-female translation.} 
    We show two translated imagee of our model under different noise vectors. 
    From left to right: input, our ACL-GAN, CycleGAN~\cite{zhu2017cyclegan}, MUNIT~\cite{huang2018munit}, StarGAN~\cite{choi2018stargan}, DRIT++~\cite{Lee2018DRIT,Lee2019DRITplus}, and CouncilGAN~\cite{Nizan2019breackcycle}.} 
    \label{fig:male}
\end{figure*}

\begin{figure*}[t]
    \begin{center}
        \includegraphics[scale=0.58, trim={4.5cm, 2.7cm, 4.5cm, 2.6cm}, clip]{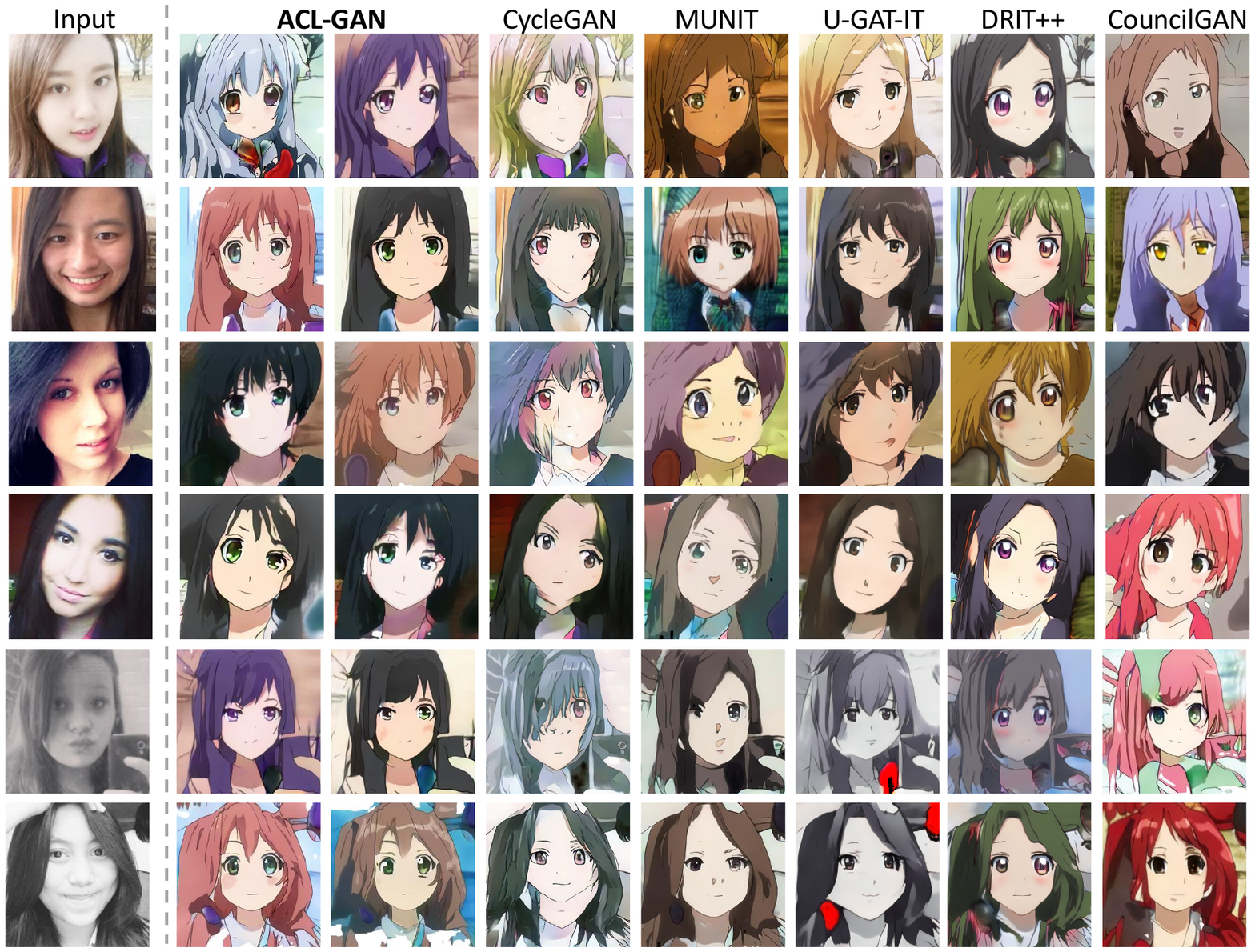}
    \end{center}
    \caption{\textbf{Comparison against baselines on selfie-to-anime translation.} 
    From left to right: input, our ACL-GAN, CycleGAN~\cite{zhu2017cyclegan}, MUNIT~\cite{huang2018munit}, U-GAT-IT~\cite{Kim2020U-GAT-IT:}, DRIT++~\cite{Lee2018DRIT,Lee2019DRITplus}, and CouncilGAN~\cite{Nizan2019breackcycle}.
    }
    \label{fig:selfie}
\end{figure*}

\subsection{Comparison with Baselines}

\begin{table}[t]
\centering
\begin{tabular}{c|cccc}
\hline
\textbf{Model}  & CycleGAN & MUNIT    & DRIT++     & Fixed-Point GAN        \\ \hline
\textbf{Params} & 28.3M    & 46.6M    & 65.0M      & 53.2M                  \\ \hline \hline
\textbf{Model}  & StarGAN  & U-GAT-IT & CouncilGAN & \textbf{ACL-GAN} \\ \hline
\textbf{Params} & 53.3M    & 134.0M   & 126.3M     & \textbf{54.9M}         \\ \hline
\end{tabular}
\caption{\textbf{The number of parameters of our method and baselines.} We set the image size to be 256 $\times$ 256 for all methods. 
The parameters of U-GAT-IT~\cite{Kim2020U-GAT-IT:} is counted in light mode. 
The number of councils is set to be four in CouncilGAN~\cite{Nizan2019breackcycle}. }
\label{table:params}
\end{table}

We compared our ACL-GAN with baselines on three challenging applications: glasses removal, male-to-female translation, and selfie-to-anime translation. 
These three applications have different commonalities between the source and target domains, and they require changing areas with different sizes. Glasses removal requires changing the smallest area, while the selfie-to-anime translation requires changing the largest area. 
Table~\ref{table:params} shows the number of parameters of our method and baselines.
Table~\ref{table:quan} shows the quantitative results for different applications and our method outperforms the baselines. 
Fig.~\ref{fig:glasses},~\ref{fig:male} and~\ref{fig:selfie} show the results of our method and baselines. 

\begin{table}[t]
\centering
\begin{tabular}{|c|c|c|c|c|c|c|c|c|}
\hline
\multirow{2}{*}{Model}                                    & \multicolumn{2}{c|}{glasses removal}                                               & \multirow{2}{*}{Model} & \multicolumn{2}{c|}{male to female}                                              & \multirow{2}{*}{Model} & \multicolumn{2}{c|}{selfie to anime}                                             \\ \cline{2-3} \cline{5-6} \cline{8-9} 
                                                          & FID            & KID                                                             &                        & FID            & KID                                                             &                        & FID            & KID                                                             \\ \hline
CycleGAN                                                  & 48.71          & \begin{tabular}[c]{@{}c@{}}0.043\\ \tiny{$\pm 0.0011$}\end{tabular}          & CycleGAN               & 21.30          & \begin{tabular}[c]{@{}c@{}}0.021\\ \tiny{$\pm 0.0003$}\end{tabular}          & CycleGAN               & 102.92         & \begin{tabular}[c]{@{}c@{}}0.042\\ \tiny{$\pm 0.0019$}\end{tabular}          \\ \hline
MUNIT           & 28.58          & \begin{tabular}[c]{@{}c@{}}0.026\\ \tiny{$\pm 0.0009$}\end{tabular}          & MUNIT                  & 19.02          & \begin{tabular}[c]{@{}c@{}}0.019\\ \tiny{$\pm 0.0004$}\end{tabular}          & MUNIT                  & 101.30         & \begin{tabular}[c]{@{}c@{}}0.043\\ \tiny{$\pm 0.0041$}\end{tabular}          \\ \hline
DRIT++                                                    & 33.06          & \begin{tabular}[c]{@{}c@{}}0.026\\ \tiny{$\pm 0.0006$}\end{tabular}          & DRIT++                 & 24.61          & \begin{tabular}[c]{@{}c@{}}0.023\\ \tiny{$\pm 0.0002$}\end{tabular}          & DRIT++                 & 104.40         & \begin{tabular}[c]{@{}c@{}}0.050\\ \tiny{$\pm 0.0028$}\end{tabular}          \\ \hline
\begin{tabular}[c]{@{}c@{}}Fixed-Point\\ GAN\end{tabular} & 44.22          & \begin{tabular}[c]{@{}c@{}}0.038\\ \tiny{$\pm 0.0009$}\end{tabular}          & StarGAN                & 36.17          & \begin{tabular}[c]{@{}c@{}}0.034\\ \tiny{$\pm 0.0005$}\end{tabular}          & U-GAT-IT               & 99.15          & \begin{tabular}[c]{@{}c@{}}0.039\\ \tiny{$\pm 0.0030$}\end{tabular}          \\ \hline
CouncilGAN                                                & 27.77          & \begin{tabular}[c]{@{}c@{}}0.025\\ \tiny{$\pm 0.0011$}\end{tabular}          & CouncilGAN             & 18.10          & \begin{tabular}[c]{@{}c@{}}0.017\\ \tiny{$\pm 0.0004$}\end{tabular}          & CouncilGAN             & 98.87          & \begin{tabular}[c]{@{}c@{}}0.042\\ \tiny{$\pm 0.0047$}\end{tabular}          \\ \hline
ACL-GAN                                                   & \textbf{23.72} & \textbf{\begin{tabular}[c]{@{}c@{}}0.020\\ \textbf{\tiny{$\pm$ 0.0010}}\end{tabular}} & ACL-GAN                & \textbf{16.63} & \textbf{\begin{tabular}[c]{@{}c@{}}0.015\\ \textbf{\tiny{$\pm$ 0.0003}}\end{tabular}} & ACL-GAN                & \textbf{93.58} & \textbf{\begin{tabular}[c]{@{}c@{}}0.037\\ \textbf{\tiny{$\pm$ 0.0036}}\end{tabular}} \\ \hline
\end{tabular}
\caption{\textbf{Quantitative results of glasses removal, male-to-female translation, and selfie-to-anime translation.} 
For KID, mean and standard deviation are listed.
A lower score means better performance. 
U-GAT-IT~\cite{Kim2020U-GAT-IT:} is in light mode.
Our method outperforms all other baselines in all applications.
}
\label{table:quan}
\end{table}

\textbf{Glasses removal}
The goal of glasses removal is to remove the glasses of a person in a given image. 
There are two difficulties in this application. First, the area outside the glasses should not be changed, which requires the generator to identify the glasses area.
Second, the glasses hide some information of the face, such as the eyes and eyebrows. The sunglasses in some images make it even more difficult because the eyes are totally occluded and the generator is expected to generate realistic and suitable eyes.

Fig.~\ref{fig:glasses} shows the results of ACL-GAN and baselines for glasses removal.
Our results leave fewer traces of glasses than~\cite{zhu2017cyclegan,huang2018munit,siddiquee2019learning,Lee2018DRIT,Lee2019DRITplus} because cycle-consistency is not required.
The left columns of Table~\ref{table:quan} show the quantitative results of glasses removal. Our method outperforms all baselines on both FID and KID.
It is interesting that our method can outperform CouncilGAN and MUNIT, which use the same network architecture as ACL-GAN.
Compared with MUNIT, our method does not require the image translated back to be a specific image so that no artefacts remain.
Compared with CouncilGAN, it might be that, rather than requiring multiple duplicate generators to compromise each other by minimising the distances between their synthesised images,
our method explicitly encourages the generated image to be similar to the source image.
 


\textbf{Male-to-female translation}
The goal of male-to-female translation is to generate a female face when given a male face, and these two faces should be similar except for gender.
Comparing with glasses removal, this task does not require to remove objects, but there are three difficulties of this translation task.
First, this task is a typical multi-modal problem because there are many possible ways to translate a male face to a female face \emph{e.g.,} different lengths of hair. 
Second, translating male to female realistically requires not only changing the colour and texture but also shape, \emph{e.g.,} the hair and beard.
Third, the paired data is impossible to acquire, \emph{i.e.,} this task can only be solved by using unpaired training data.

Fig.~\ref{fig:male} compares our ACL-GAN with baselines for male-to-female translation. 
For each row, two images of ACL-GAN are generated with two random noise vectors. This shows the diversity of the results of our method, which is the same in Fig.~\ref{fig:selfie}.
Our results are more feminine than baselines with cycle-consistency loss~\cite{zhu2017cyclegan,choi2018stargan,Lee2018DRIT,Lee2019DRITplus} as our faces have no beard, longer hair and more feminine lips and eyes.
We found that MUNIT~\cite{huang2018munit} can generate images with long hair and no beard, which may contribute to the style latent code.
However, the important features, $\emph{e.g.}$ the hue of the image, cannot be well-preserved.
Quantitatively, the middle columns of Table~\ref{table:quan} show the effectiveness of our method and it can be also explained by the lack of cycle-consistency loss~\cite{zhu2017cyclegan} and compromise between duplicated generators~\cite{Nizan2019breackcycle}.

\textbf{Selfie-to-anime translation} 
Different from the previous two tasks, 
generating an animated image conditioned on a selfie
requires large modifications of shape.
The structure and style of the selfie are changed greatly in the target domain, \emph{e.g.,} the eyes become larger and the mouth becomes smaller.
This may lead to the contortion and dislocation of the facial features.

Fig.~\ref{fig:selfie} shows the results of ACL-GAN and baselines for selfie-to-anime translation.
The results of methods with cycle-consistency are less in accordance with the style of anime. 
In contrast, without cycle-consistency loss, our method can generate images more like an anime, \emph{e.g.} the size and the layout of facial features are better-organized than baselines. The adversarial-consistency loss helps to preserve features, \emph{e.g.,} the haircut and face rotation of the source images.
U-GAT-IT~\cite{Kim2020U-GAT-IT:} utilises an attention module and AdaIN to produce more visually pleasing results.
The full mode of U-GAT-IT has $670.8$M parameters, which is more than twelve times the number of parameters our model has.
For fairness, we evaluated U-GAT-IT in light mode for comparison.
Our method can still outperform U-GAT-IT when it is in light mode, which still uses more than twice as many parameters as ours as shown in Table~\ref{table:params}.

\begin{figure}[t]
    \begin{center}
        \includegraphics[scale=0.6, trim={7cm, 7.8cm, 7cm, 7.8cm}, clip]{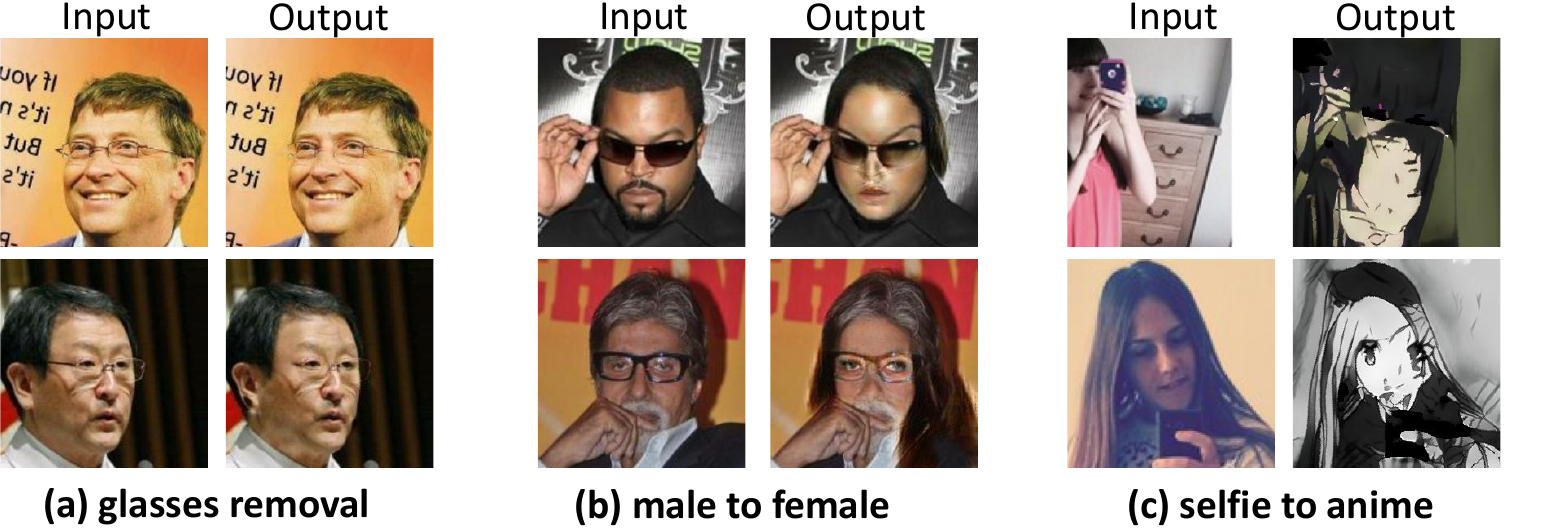}
    \end{center}
    \caption{\textbf{Typical failure cases of our method.}
    (a) Some glasses are too inconspicuous to be identified and they can not be completely removed by the generator.
    (b) We also found that the glasses might be partly erased when translating a male to a female and this is because of the imbalance inherent to the datasets.
    (c) The generated anime image may be distorted and blurred when the face in the input image is obscured or too small, because it is far from the main distribution of the selfie domain.
    }
    \label{fig:failure}
\end{figure}

\section{Limitations and Discussion} 
In this paper, we present a novel framework, ACL-GAN, for unpaired image-to-image translation.
The core of ACL-GAN, adversarial-consistency loss, helps to maintain commonalities between the source and target domains.
Furthermore, 
ACL-GAN can perform shape modifications and remove large objects without unrealistic traces.

Although our method outperforms state-of-the-art methods both quantitatively and qualitatively on three challenging tasks, typical failure cases are shown in Fig.~\ref{fig:failure}. 
We also tried learning to translate real images from domain $X_S$ to $X_T$ and from domain $X_T$ to $X_S$ synchronously, but the performance was decreased, which may be caused by the increase of network workload.
In addition, our method is not yet able to deal with datasets that have complex backgrounds (\emph{e.g.} horse-to-zebra) well.
Thus, supporting images with a complex background is an interesting direction for future studies.
Nevertheless, the proposed method is simple, effective, and we believe our method can be applied to different data modalities.

\section{Acknowledgements}
This work was supported by the funding from Key-Area Research and Development Program of Guangdong Province (No.2019B121204008), start-up research funds from Peking University (7100602564) and the Center on Frontiers of Computing Studies (7100602567). We would also like to thank Imperial Institute of Advanced Technology for GPU supports.

\clearpage
%
%
\bibliographystyle{splncs04}
\bibliography{egbib}
\end{document}